%% file: main.tex
\documentclass[10pt,twocolumn,letterpaper]{article}

\usepackage[pagenumbers]{iccv} 
\usepackage{tikz}
\usepackage{pgfplots}
\pgfplotsset{compat=default}
\usetikzlibrary{pgfplots.groupplots}
\usepackage{longtable}
\usepackage{xcolor}
\usepackage{multirow}

\definecolor{iccvblue}{rgb}{0.21,0.49,0.74}
\usepackage[pagebackref,breaklinks,colorlinks,allcolors=iccvblue]{hyperref}

\def\paperID{****} 
\def\confName{ICCV}
\def\confYear{2025}

\title{The Power of One: A Single Example is All it Takes for Segmentation in VLMs}

\author{Mir Rayat Imtiaz Hossain$^{1,2}$ \qquad Mennatullah Siam$^{1}$  \qquad Leonid Sigal$^{1,2,3}$ \qquad James J. Little$^{1}$\\
$^1$University of British Columbia \qquad $^2$Vector Institute for AI  \qquad $^3$Canada CIFAR AI Chair \\}
\begin{document}
\maketitle
\input{sec/0_abstract}    
\input{sec/1_intro}
\input{sec/2_relatedWork}

\input{sec/3_method}

\input{sec/4_results}
\input{sec/5_conc}
\footnotetext{ \textbf{Acknowledgments and Disclosure of Funding.} This work was partially funded by the Vector Institute for AI, Canada CIFAR AI Chair, NSERC CRC and NSERC DG. Hardware resources were partially provided by the Province of Ontario, the Government of Canada through CIFAR, and \href{https://vectorinstitute.ai/\#partners}{companies} sponsoring the Vector Institute. Additional support was provided by JELF CFI grant and Digital Research Alliance of Canada under the RAC award. }

{ \small \bibliographystyle{ieeenat_fullname} 

\input{main.bbl}
}

\input{sec/X_suppl}


\end{document}

%% file: sec/0_abstract.tex
\begin{abstract}
Large-scale vision-language models (VLMs), trained on extensive datasets of image-text pairs, exhibit strong multimodal understanding capabilities by implicitly learning associations between textual descriptions and image regions. This emergent ability enables zero-shot object detection and segmentation, using techniques that rely on text-image attention maps, without necessarily training on abundant labeled segmentation datasets. However, performance of such methods depends heavily on prompt engineering and manually selected layers or head choices for the attention layers. In this work, we demonstrate that, rather than relying solely on textual prompts, providing a single visual example for each category and fine-tuning the text-to-image attention layers and embeddings significantly improves the performance. Additionally, we propose learning an ensemble through few-shot fine-tuning across multiple layers and/or prompts. An entropy-based ranking and selection mechanism for text-to-image attention layers is proposed to identify the top-performing layers without the need for segmentation labels. This eliminates the need for hyper-parameter selection of text-to-image attention layers, providing a more flexible and scalable solution for open-vocabulary segmentation. We show that this approach yields strong zero-shot performance, further enhanced through fine-tuning with a single visual example. Moreover, we demonstrate that our method and findings are general and can be applied across various vision-language models (VLMs).  
\end{abstract}

%% file: sec/1_intro.tex
\section{Introduction}
\label{sec:intro}

\begin{figure}[h]
\centering
    \includegraphics[width=0.42\textwidth]{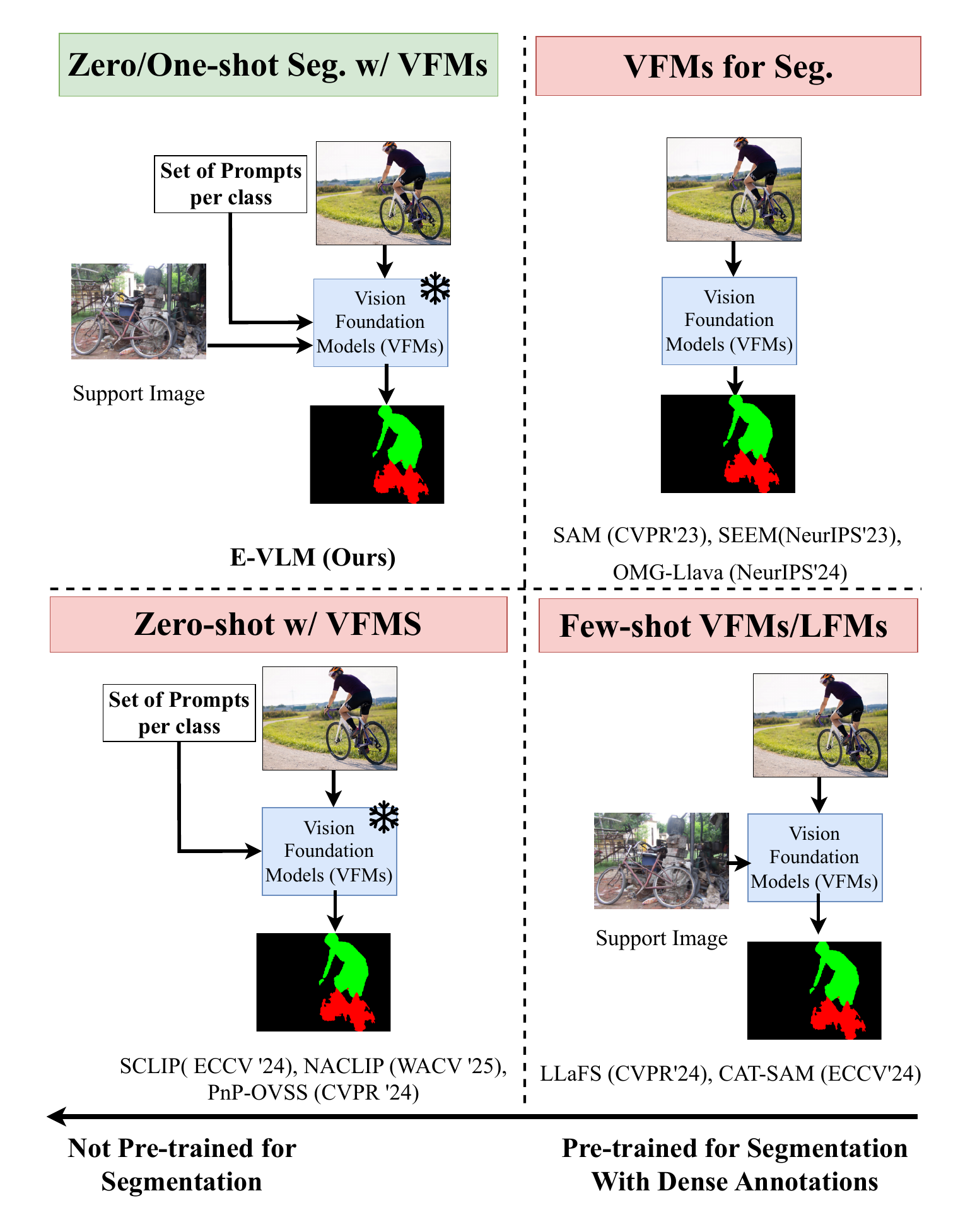}
    \vspace{-0.1in}
    \caption{{\bf Segmentation using vision foundational models (VFMs)} can be broadly categorized based on their pre-training: the right half includes models pretrained specifically for segmentation with dense annotations, while the left half comprises models not pretrained for segmentation tasks. Each category is further divided into four distinct approaches (clockwise): the top right contains models trained on extensive data for segmentation; the bottom right includes models pretrained for segmentation then evaluated on novel categories with few-shot data; the bottom left represents training-free segmentation models, typically vision-language models (VLMs) trained solely on image-text pairs; and the top left (ours) features a hybrid approach allowing both training-free inference and one-shot fine-tuning.} 
    \label{fig:overview}
    \vspace{-1em}
\end{figure}

In recent years, deep learning research has shifted towards foundation models~\cite{bommasani2021opportunities, li2024multimodal}, which are trained on broad datasets to support generalization across a wide variety of downstream tasks, primarily using self-supervised learning~\cite{he2022masked, oquab2023dinov2, caron2021emerging} or vision-language modeling~\cite{radford2021learning, xu2023bridgetower, li2022blip,li2023blip,alayrac2022flamingo}. Vision-language pre-training, in particular, has seen significant advancements with models like CLIP~\cite{radford2021learning} and BLIP~\cite{li2022blip}, which leverage image-text contrastive learning or image-text matching. Following these foundational models, multi-modal large language models (MLLMs) have emerged—such as Flamingo~\cite{alayrac2022flamingo} and LLaVA~\cite{liu2024visual}—building on the capabilities of large language models (LLMs) that excel in natural language tasks. Further, new foundation models like the Segment Anything Model (SAM)~\cite{kirillov2023segment, ravi2024sam} and LLaVA descendants~\cite{chen2023llava,zhang2024omg} focus on pixel-level or region-level understanding, though they require extensive datasets with pixel-level annotations.

There is a concurrent stream of research in pixel-level understanding tasks, particularly in image segmentation, that focuses on training-free methods. These approaches leverage information from vision-language models for segmentation tasks and are evaluated in the open-vocabulary segmentation setting~\cite{wang2025sclip, hajimiri2024pay, luo2024emergent, zhou2022extract,cha2023learning,barsellotti2024fossil}. However, to the best of our knowledge, the potential of using few-shot demonstrations has not been explored in these models. While open-vocabulary segmentation is typically focused on the zero-shot setting, we demonstrate that incorporating a single visual example within a carefully designed framework can yield substantial improvements in performance. 

 In this paper, we explore the potential of VLMs for image segmentation, leveraging the power of a single demonstration, as illustrated in Fig.~\ref{fig:overview}. We propose two key components in our approach that enable its extensibility across various VLMs without requiring extensive hyperparameter tuning for layer selection, heads, or prompts. Our method employs an entropy-based, automatic selection and ranking of text-to-image attention layers in an unsupervised manner, along with re-weighting these attention maps using an image-text scoring mechanism. Furthermore, we introduce a novel ensemble learning during few-shot fine-tuning that combines text-to-image attention maps from multiple layers and/or prompts. Our approach demonstrates significant performance improvements across four benchmarks and strong extensibility across VLMs including the earlier variants ({\em i.e.}, ALBEF~\cite{li2021align} and BLIP~\cite{li2022blip}) and recent LLM based ones ({\em i.e.}, LLaVA 1.5~\cite{liu2024improved}). It operates in two modes; training-free and one-shot fine-tuning, where we show our training-free approach outperforms previous works and is considerably improved with one-shot fine-tuning.

\noindent
{\bf Contributions.} 
In summary, our contributions include:
\begin{itemize}
\item The design of a few-shot fine-tuning mechanism for vision-language models (VLMs) to extract image segmentation without the need to exhaustively train VLMs on abundant image segmentation datasets or adding additional decoders or parameters.
\item At the core of our design are two novel mechanisms: \textbf{(i)} using an entropy-based metric to rank text-to-image attention layers in an unsupervised manner; \textbf{(ii)} re-weighting the attention maps using image-text scoring.
\item We propose a novel ensemble learning mechanism that relies on fine-tuning the attention maps extracted from multiple layers and/or prompts with the one visual example.
\end{itemize}

%% file: sec/2_relatedWork.tex
\section{Related Work}

\subsection{Open Vocabulary Segmentation} 

Zero-shot semantic segmentation has been extensively studied to segment unseen, novel classes using text descriptions~\cite{bucher2019zero, zhou2023zegclip}. However, these methods do not allow overlap between base and novel classes. Open-vocabulary segmentation addresses this by allowing such overlap and leveraging large-scale vision-language pre-training~\cite{li2022language, wu2024towards}. Training with language data allows the models to handle large, extensible vocabularies, unlike traditional zero-shot settings with predefined base categories. Recent advances in vision-language pre-training~\cite{radford2021learning, li2022blip} have further advanced open-vocabulary image segmentation.

Recent open-vocabulary segmentation methods have focused on training-free approaches to mine segmentation masks from vision-language pre-trained models. PnP-OVSS~\cite{luo2024emergent} introduced a plug-and-play method that uses cross-attention maps from VLMs and saliency dropout for refinement, while requiring an external model to filter out irrelevant classes. SCLIP~\cite{wang2025sclip} adapted CLIP by replacing self-attention with correlative self-attention, and NACLIP~\cite{hajimiri2024pay} utilized self-attention in CLIP to enforce local spatial consistency. However, these methods do not leverage few-shot or one-shot demonstrations, making them susceptible 
to distributions of data and vocabularies for which original VLMs have been trained. 
In contrast, our work extends open-vocabulary segmentation by incorporating such demonstrations to mine VLMs for image segmentation.

\subsection{Few-shot Segmentation} 

Few-shot segmentation has been extensively studied in prior works, focusing primarily on traditional approaches that do not leverage recent advances in vision-language models~\cite{wang2019panet,siam2019amp,shaban2017one,min2021hypercorrelation}. These approaches typically address 1-way or N-way segmentation of novel classes against a background but do not evaluate performance on the base classes seen during pre-training. Recent studies have proposed a generalized few-shot segmentation setting that enables evaluation on both base and novel classes~\cite{tian2022generalized,hajimiri2023strong,hossain2024visual, liu2023learning}. A recent approach~\cite{hossain2024visual} uses a multiscale visual prompting technique to improve segmentation. However, none of the works in this category fully utilize the capabilities of vision foundation models, particularly VLMs.

\begin{figure*}[t!]
    \includegraphics[width=0.95\textwidth]{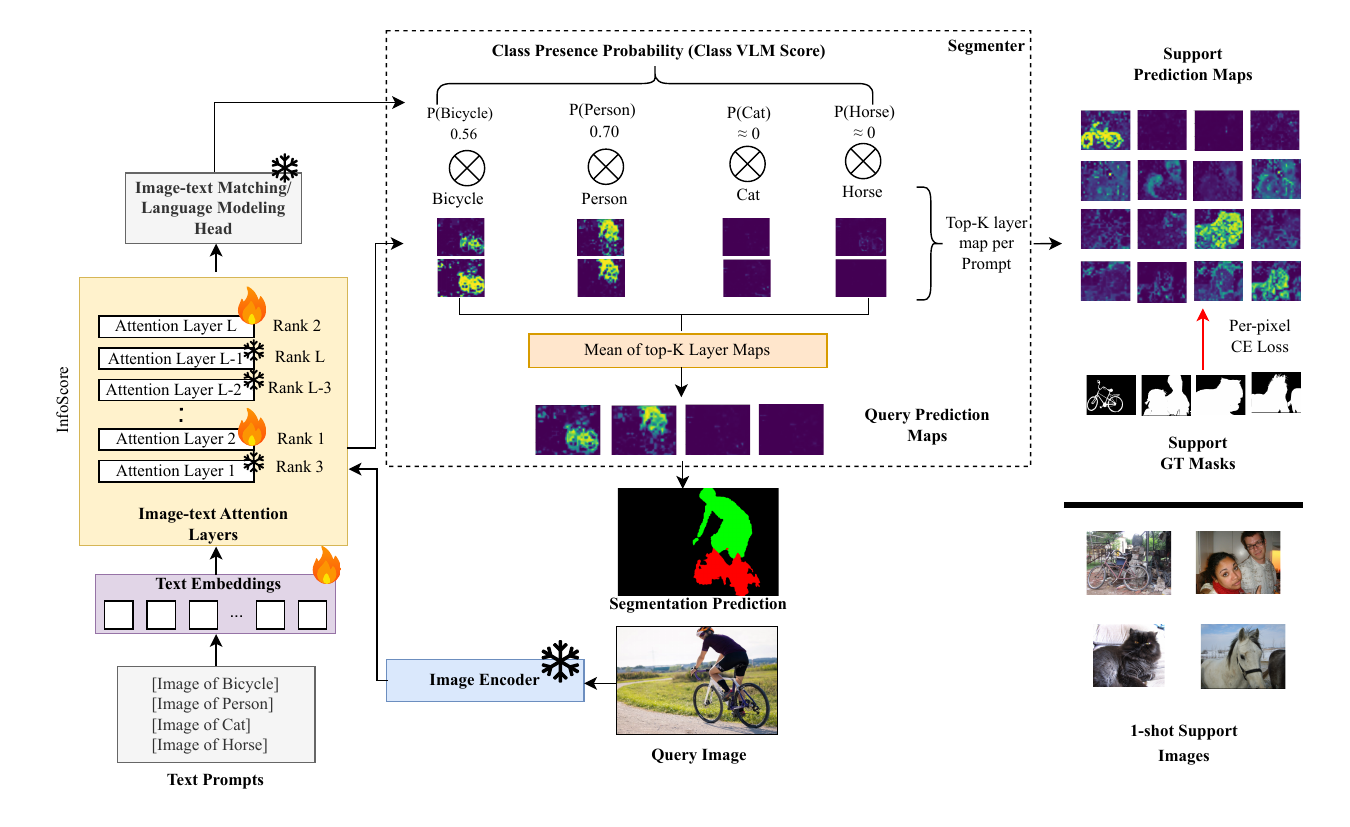}
    \vspace{-0.2in}
    \caption{{\bf Model Overview.} Our segmentation framework leverages VLMs trained on image-text pairs, supporting training-free inference and one-shot fine-tuning. For training-free inference, given class names and a query image, we extract text-to-image attention maps from top-$K$ layers (e.g., Layer 2 and Layer \textit{L}), selected via InfoScore (see Sec.~\ref{subsec:infoscore}). These maps are re-weighted with class VLM scores to filter irrelevant categories (see Sec.~\ref{subsec:ca_map_description}) and used for prediction. In one-shot fine-tuning, we adjust text embeddings and top-$K$ attention layer parameters (see Sec.~\ref{subsec:fine-tune}) to further improve the performance.}
    \label{fig:model_architecture}
\end{figure*}


A concurrent work~\cite{catsam} has explored few-shot adaptation of the the recent segment anything (SAM) model~\cite{kirillov2023segment}. However, SAM is already trained with dense mask annotations over millions of data-point. In contrast, we utilize VLMs that were only trained with image-text pairs, and not trained for segmentation. A recent study \cite{zhu2024llafs} explored few-shot segmentation with LLMs and an additional mask decoder by instruction-tuning the LLM on pixel-level annotations to generate 16-point polygons, while also training the mask decoder on a large dataset of pixel-level annotations for base categories.
Moreover, their method only supports simple 1-way segmentation (segmenting foreground class from background). In contrast, our approach directly mines multi-class segmentation maps from VLM text-to-image attention maps without relying on external models like ChatGPT and requires no additional training. Furthermore, fine-tuning just a few parameters with a single visual example leads to significant performance gains.

%% file: sec/3_method.tex
\section{Methodology}

Our proposed approach leverages text-to-image attention maps from the VLMs. We specifically focus on VLMs that are pre-trained on large datasets of image-text pairs without segmentation annotations except for using only a single visual example, {\em i.e.}, one-shot setting. Figure~\ref{fig:model_architecture} presents an overview of our proposed approach, which operates in two modes: either in a training-free mode or fine-tuned with one visual example. In the training-free setting, we provide a set of text prompts along with a query image. We then extract attention maps from the top-$K$ text-to-image attention layers, ranked by our proposed InfoScore metric, and re-weight them using image-text scores; see Sec.~\ref{subsec:ca_map_description} and~\ref{subsec:infoscore} respectively. In the one-shot scenario, we leverage one visual example for each category, with corresponding segmentation masks, to learn an ensemble from multiple layers and/or prompts, see Sec.~\ref{subsec:fine-tune}.

\subsection{Preliminaries}
\textbf{Vision-language Pretraining. } This work focuses on a class of vision-language models (VLMs) that use image-text cross-attention for multi-modal reasoning, specifically we use BLIP~\cite{li2022blip} and ALBEF~\cite{li2021align}, but are not restricted to these. Our method is also extensible to the recent MLLMs such as LLaVA 1.5~\cite{liu2024improved}.

BLIP \cite{li2022blip} employs jointly trained image and text encoders with a contrastive image-text loss to learn aligned representations. These aligned representations are then processed by a multi-modal encoder that is optimized with an image-text matching (ITM) loss to reinforce cross-modal alignment. Similarly, ALBEF\cite{li2021align} uses a pre-training phase where image and text encoders are aligned through contrastive losses on image-text pairs. A multi-modal encoder is subsequently trained with both ITM loss and masked language modeling (MLM) loss, making it adaptable to various downstream tasks.
In both BLIP and ALBEF, text-to-image attention leverages cross-attention layers. Recent MLLM models like LLaVA~\cite{liu2024visual,liu2024improved}, on the other hand, rely on causal self-attention between visual and textual tokens. In such cases, a portion of self-attention matrix effectively serves the same role as the cross-attention.

\vspace{0.05in}
\noindent
\textbf{Training-free Approaches.}
Although some VLMs have not been specifically trained for segmentation, like SAM~\cite{kirillov2023segment} or SEEM~\cite{zou2024segment}, it has been previously demonstrated that image-text cross-attention maps can effectively facilitate zero-shot segmentation~\cite{luo2024emergent}. However, ~\cite{luo2024emergent} also observed that cross-attention maps tend to over-segment objects and generate numerous false positives, and therefore used GPT-4o to filter out categories not present in the image. We observe that we can refine the heatmaps using an image-text scoring that come with these VLMs, without the need for any additional models beyond the VLM itself.

Another common observation in approaches that utilize image-text cross-attention maps, or Grad-CAM, for tasks such as segmentation~\cite{luo2024emergent}, visual grounding~\cite{he2024improved}, or visual question answering~\cite{tiong2022plug} is that performance often depends on the specific layer or head from which cross-attention maps are extracted. Typically, these methods assess layer-wise, task-specific performance on validation sets with ground-truth annotations to select the layer that maximizes performance~\cite{tiong2022plug}, or they use ground-truth class information from validation images~\cite{luo2024emergent}. Both approaches, however, rely on some form of ground truth, which limits their practicality in a truly training-free, open-vocabulary setting where annotations are unavailable, making it challenging to determine which layer’s cross-attention map would yield strong segmentation performance. To address this limitation, we propose an entropy-based metric called InfoScore, designed to identify the most suitable layer combinations in a vision-language model (VLM), without annotations.

\subsection{Computing Class-wise Heatmaps}
\label{subsec:ca_map_description}

As shown in Figure~\ref{fig:model_architecture}, during inference, we pass a list of prompts in the format \texttt{\small [[Image of Class 1], [Image of Class 2], ... [Image of Class N]]} along with a query image divided into $P \times P$ patches into the VLM. At each cross-attention layer of the multimodal encoder, each text prompt interacts with the image tokens (patches).
Given $N$ text prompts (representing $N$ classes) and the maximum length of a prompt being $T$, for each text prompt and cross-attention layer, we obtain an attention score of dimension $T \times P \times P$. Since each cross-attention layer has multiple heads, we compute the maximum attention score across all heads for each text prompt.

To compute the overall attention score for a prompt, we take the mean across the words in the prompt, resulting in an attention score of dimension $P \times P$ for that prompt. This process is repeated for each class prompt in the prompt list, and the scores are concatenated, yielding an overall attention score tensor of dimension $N \times P \times P$. Finally, this tensor is normalized by applying \texttt{softmax} across the $N$ classes to obtain a per-patch probability estimate for each class. In cases where we use related words to enhance the class description, similar to previous works~\cite{hajimiri2024pay, wang2025sclip}, {\em e.g.}, \texttt{man} and \texttt{woman} added as related words to the class \texttt{person}, we pass them as separate prompts. To compute the heatmap for the corresponding class, we simply take the maximum across the heatmaps of all the related prompts. We call this the {\em multi-prompt} setting.

Once per-class attention maps are obtained, we select the attention maps from the top-$K$ layers ranked by our InfoScore metric, discussed in Section~\ref{subsec:infoscore}. The InfoScore metric ranks attention maps from different layers in an unsupervised manner based on their predictive uncertainty. Attention maps from transformer-based models and VLMs tend to be noisy due to attention sinks, particularly in models leveraging large language models (LLMs) like LLaVA-1.5~\cite{liu2024improved}, as noted in prior works~\cite{kangsee,registervit,sun2024massive}. This issue, combined with our approach of aggregating attention maps across different prompts, frequently results in high attention responses for classes not present in the image, leading to false positive predictions.

Since many VLMs are trained with an image-text matching (ITM) loss, we observe that given a prompt such as \texttt{\small [Image of class N.]}, the ITM head provides a probability indicating how well the prompt describes the image, assigning higher probabilities when the class is present and lower probabilities otherwise. For models not trained with ITM loss and lacking an ITM head ({\em e.g.}, LLaVA-1.5), we instead prompt them with \texttt{\small [Is there class N in the image? Answer in Yes or No.]} and estimate the probability of \texttt{\small class N} being present based on the probability of the next token prediction being \texttt{\small [Yes]}. We define this probability-based scoring of the presence of a class as \textbf{class VLM score}. The class VLM scores are then multiplied with the image-text attention maps from the selected layers, effectively suppressing noisy attention responses for absent classes while amplifying those corresponding to present classes. Finally, we aggregate the filtered attention maps from the top-$K$ layers using mean. The ensembled maps are subsequently refined using a convolutional conditional random field (ConvCRF)~\cite{teichmann2018convolutional} for better boundary delineation.


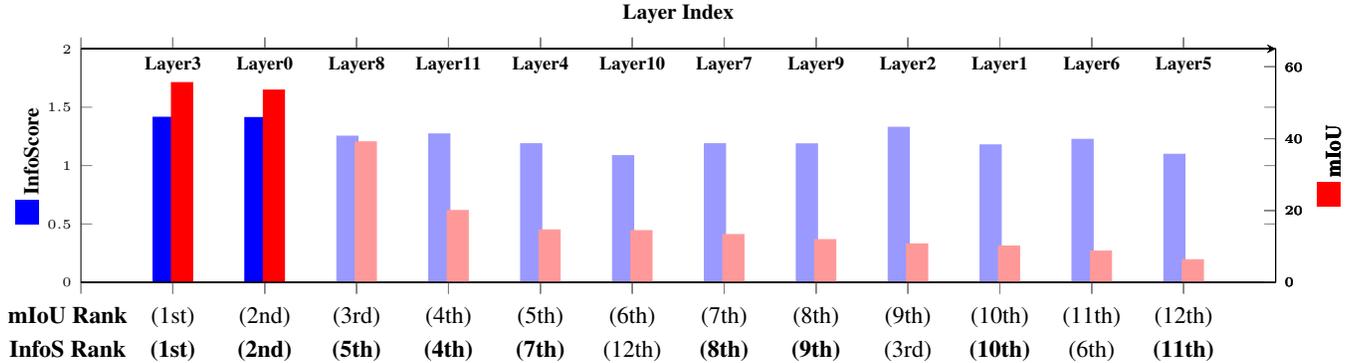
\begin{figure*}[ht]
\centering
\input{Images/plot_infoscore_v2}
\caption{{\bf Illustration of the InfoScore Metric on BLIP.} The mIoU Rank reflects the descending order of mIoU values (in red) derived from cross-attention maps for each standalone layer (labeled Layer$N$, top) on the PASCAL VOC 2012 validation set (1449 images), compared to the predicted InfoS Rank (bottom) based on our InfoScore metric (in blue) requiring no annotations. Most InfoScore rankings align with mIoU Rankings, with minor displacements of $\pm2$ positions highlighted in bold, except for four layers. Empirically, the top-1 and top-2 layers are correctly identified and consistently deliver better performance across four datasets and three VLMs.}
\label{tab:layer_ranking_table}
\end{figure*}

\subsection{Ranking Layers Using InfoScore}
\label{subsec:infoscore}

As noted in previous works~\cite{luo2024emergent, he2024improved, tiong2022plug}, and as observed in Figure~\ref{tab:layer_ranking_table}, segmentation performance can vary significantly depending on the layer from which the attention map is extracted. In practice, when performing training-free segmentation in-the-wild, ground-truth dense annotations or information about object presence in the images is unavailable. To address this, we propose an entropy-based metric, InfoScore, which can identify the optimal layer(s) for training-free segmentation without ground-truth annotations.

Given a set of unlabelled images, $\mathcal{D}$, and a vocabulary of class names, the proposed InfoScore metric evaluates the predictive uncertainty of text-to-image attention maps from each layer individually. The InfoScore metric consists of two key components: (i) \textit{mean image-level entropy}, defined as the average per-image entropy of the label marginal distribution summed over the pixels, and (ii) \textit{dataset-level entropy}, defined as the entropy of the class-wise marginal distribution across the entire dataset; motivated by the Inception Score~\cite{salimans2016improved}. 

Given the per-pixel probability predictions for an image $\mathbf{I}$, we compute the image-level classwise marginal distribution using layer $l$ cross-attention maps, $P^l_I(\hat{\mathbf{Y}_c})$ as,

\begin{equation}
P^l_I(\hat{\mathbf{Y}}_c) = \frac{1}{hw} \sum_{i=1}^{h} \sum_{j=1}^{w} P^l_{I_{x,y}}(\hat{\mathbf{Y}}_c),
\end{equation}

\noindent where $h, w$ are the height and width of the predicted heatmap, $P^l_{I_{x,y}}(\hat{\mathbf{Y}}_c)$ is the probability that the pixel $(x,y)$ belongs to class $c$ which is computed based on the cross-attention maps from layer $l$. Then we compute the image-level entropy $\mathcal{H}^l(\mathbf{I})$ as,
\begin{equation}
\mathcal{H}^l(\mathbf{I}) = - \sum_{\text{c} \in V} P^l_I(\hat{\mathbf{Y}}_c) \log P^l_I(\hat{\mathbf{Y}}_c),
\end{equation}
where $V$ is the vocabulary of the classes to segment. To compute the mean image-level entropy, we take the average over the set of unlabeled images in dataset $\mathcal{D}$,

\begin{equation}
\mathcal{H}^l_{\text{image}}(\mathcal{D}) = \frac{1}{|\mathcal{D}|} \sum_{I \in \mathcal{D}} \mathcal{H}^l(\mathbf{I}).
\end{equation}

In order to compute the dataset-level entropy over the dataset $\mathcal{D}$, we first calculate the classwise marginal distribution over the entire dataset, $P^l(\hat{\mathbf{Y}_c})$, as,

\begin{equation}
P^l(\hat{\mathbf{Y}}_c) = \frac{1}{|\mathcal{D}|} \sum_{ \mathbf{I} \in \mathcal{D}} P^l_I(\hat{\mathbf{Y}}_c)
\end{equation}

The dataset-level entropy $\mathcal{H}_{\text{dataset}}$ is then given by:

\begin{equation}
\mathcal{H}^l_{\text{dataset}}(\mathcal{D}) = - \sum_{\text{c} \in V} P^l(\hat{\mathbf{Y}}_c) \log P^l(\hat{\mathbf{Y}}_c)
\end{equation}

Finally, the InfoScore for a layer $l$ is defined as:

\begin{equation}
\text{InfoScore}(l) = \frac{\mathcal{H}^l_{\text{dataset}}(\mathcal{D}) }{\mathcal{H}^l_{\text{image}}(\mathcal{D})}
\end{equation}

To understand the motivation behind using this metric, we can view this metric as an assessment of the classification ability of each text-to-image attention layer in a self-supervised way. On the image-level, we want the classifier to accurately predict the correct classes without false positives resulting in a sharply peaked class distribution for only a few classes, {\em i.e.}, minimizing the mean image-level entropy. However, this image-level entropy can be trivially minimized by the classifier predicting the same class across all images ({\em e.g.}, if the classifier is biased towards background).

Ideally, we want the classifier to predict a diverse set of classes across images. Assuming independent and identically distributed sample images within the dataset where each class is equally likely to appear, we can aim for a uniform dataset-level label marginal distribution. Hence, while minimizing the image-level entropy we aim to maximize the dataset-level entropy. Therefore, the InfoScore is defined as the ratio of dataset-level entropy to image-level entropy. The layer that achieves the highest InfoScore is thus considered the most reliable for the use of its text-to-image attention.

\begin{table*}[ht]
\centering
\resizebox{0.90\textwidth}{!}{
\renewcommand{\arraystretch}{1.2} 
\setlength{\tabcolsep}{16pt} 
\begin{tabular}{lcccccc}
\toprule
\textbf{Method} & \textbf{VLM} & \textbf{PASCAL-21} & \textbf{COCO-Obj} & \textbf{COCO-171} & \textbf{ADE-20K} \\
\midrule

\multicolumn{6}{l}{\emph{Weakly Supervised Training}} \\
\midrule
GroupVIT~\cite{xu2022groupvit} & CLIP &  52.3 & 27.5 & 15.3 & 10.4\\
ReCo~\cite{cha2023learning} & CLIP & 51.2 & 30.4 & 19.6 & 11.2 \\
GroundEverything~\cite{bousselham2024grounding} & MetaCLIP & 46.8 & - & - & 17.1 \\
SAM-CLIP~\cite{wang2024sam} & CLIP & 60.6 & 31.5 & - & 17.1 \\
Clip-DINOiser~\cite{wysoczanska2024clip} & CLIP & 62.1 & 34.8 & 24.6 & 20.0 \\
\midrule
\multicolumn{6}{l}{\emph{Training Free}} \\
\midrule

PNP-OVSS~\cite{luo2024emergent} & BLIP & 51.3 & 36.2 & 17.9 & 14.2 \\
SCLIP~\cite{wang2025sclip} & CLIP & 61.7 & 33.2 & 23.9 & 17.8 \\
ProxyCLIP~\cite{lan2025proxyclip} & CLIP+DINO & 61.3 & 37.5 & 26.5 & 20.2 \\
NACLIP~\cite{hajimiri2024pay} & CLIP &  \textbf{64.1} & 36.2 & 25.7 & 19.1 \\
\textbf{Ours} & BLIP &  60.2 & \textbf{42.8} & \textbf{28.1} & \textbf{20.9} \\

\midrule
\multicolumn{6}{l}{\emph{One-shot Supervision} (Average Across 5 runs)} \\
\midrule
\textbf{Ours} & BLIP &  \textbf{70.1 $\pm$ 0.89} & \textbf{45.3 $\pm$ 1.32} & \textbf{29.3 $\pm$ 0.05} & \textbf{22.5$\pm$ 1.13}\\
\bottomrule
\end{tabular}
}
\caption{{\bf Comparison of our E-VLM approach with state-of-the-art methods} for weakly-supervised, training-free, and one-shot supervised open-vocabulary semantic segmentation (OVSS). The results reported on this table are using \textbf{BLIP top-k layers (selected by InfoScore)} in \textbf{multi-prompt setting}. Best results for training-free setting and the 1-shot results are in \textbf{bold}.}
\label{tab:comparison_ev}
\end{table*}

As shown in Figure~\ref{tab:layer_ranking_table}, our proposed metric successfully ranks the top performing layers on PASCAL-21. It shows our InfoScore (blue) and its corresponding rankings across 12 layers of BLIP computed over all the images in PASCAL-21 validation set. It also displays the mean intersection over union (mIoU) scores (red) for output segmentations from each layer with the ranking based on these scores. Although three layers, ranked 6th, 9th, and 11th by the oracle, were incorrectly ranked by InfoScore, the majority of layers were correctly ranked within a slight margin of error. Empirically we observe that the top-2 ranked layers 
were sufficient to drive considerable gains across multiple vision-language models and three benchmarks, see Sec.~\ref{sec:ablation}.
\vspace{-4pt}
\subsection{One-shot fine-tuning}
\label{subsec:fine-tune} 
So far, we have discussed training-free inference using standard VLMs. However, we argue that class names alone are not sufficient to properly ground or segment the corresponding objects. They can be ambiguous or confusing without the appropriate context. For example, the COCO-Obj dataset includes a class name \texttt{tie}, which can be ambiguous unless additional context is provided to indicate it refers to a piece of clothing. Therefore, we hypothesize that fine-tuning certain parameters of a VLM pretrained on an image-text retrieval task, using one visual example with dense annotations, can improve the overall segmentation quality.

In most few-shot segmentation tasks, whether performing 1-way or k-way classification, classes are typically divided into non-overlapping splits of base and novel categories. In contrast, our approach does not involve base classes, as the models we use were never explicitly trained for segmentation, meaning all classes are novel. To ensure fairness, only one novel category is present in each selected image. However, in practice, certain novel classes may co-occur. If a one-shot example image for one class contains other classes, we assign them to either the background (if the dataset includes a background category) or to an ignored class. For each example, we optimize the word embeddings and the parameters of the top-$K$ ranked attention layers using a per-pixel cross-entropy loss. It is to be noted that we do not introduce any additional parameters or decoders for fine-tuning, merely forcing the attention-maps to be close to the support image ground truth.

%% file: Images/plot_infoscore_v2.tex
\begin{tikzpicture}
\begin{axis} [
     title={},
     width=\textwidth,
     height=.208\textheight,
     ylabel={\footnotesize \tikz \fill[blue] (0,0) rectangle (0.02,0.1); \textbf{InfoScore}},
     bar width = 8pt,
     ybar = 2cm,
     xmin=0.0, xmax=13,
     xtick={1,2,3,4,5,6,7,8,9,10,11,12},
     xticklabel style = {align=center},
     xticklabels={}, 
     ymin=0.0, ymax=2,
     x tick label style={font=\scriptsize},
     y tick label style={font=\tiny},
     y label style={at={(axis description cs:0.05,.5)},anchor=south},
     x label style={at={(axis description cs:0.5,-.15)},anchor=south},
     ymajorgrids=false,
     xmajorgrids=false,
] 
\addplot[color=blue!40, fill=blue!40] coordinates {(0.9, 1.413) (1.9, 1.411) (2.9, 1.252) (3.9, 1.271) (4.9, 1.187) (5.9, 1.085) (6.9, 1.187) (7.9, 1.186) (8.9, 1.328) (9.9, 1.177) (10.9, 1.225) (11.9, 1.097) };
\end{axis}

\begin{axis} [
     title={},
     width=\textwidth,
     height=.208\textheight,
     axis x line=none,
     axis y line=none,
     major x tick style = transparent,
     major y tick style = transparent,
     bar width = 8pt,
     ybar = .05cm,
     xmin=0.0, xmax=13,
     ymin=0.0, ymax=2,
     ymajorgrids=false,
     xmajorgrids=false,
] 
\addplot[color=blue, fill=blue] coordinates {(0.9, 1.413) (1.9, 1.411)};
\end{axis}

\begin{axis}[
    axis y line=none,
    axis x line=top,
    width=\textwidth,
    height=.208\textheight,
    xlabel style={at={(axis description cs:0.5,0.98)},anchor=south,font=\footnotesize},
    xtick=\empty,
    ytick=\empty,
    xmin=0.0, xmax=13,
    ymin=0.0, ymax=2,
]
\end{axis}

\begin{axis}[
    axis y line*=right,
    width=\textwidth,
    height=.208\textheight,
    bar width=8pt,
    ybar = .05cm,
    ylabel = {\footnotesize \tikz \fill[red] (0,0) rectangle (0.02,0.1); \textbf{mIoU}},
    y tick label style={font=\tiny},
    x tick label style={yshift=0.80*\pgfkeysvalueof{/pgfplots/height}, font=\scriptsize},
    y label style={at={(axis description cs:1.14,.5)},anchor=south},
    xtick={1,2,3,4,5,6,7,8,9,10,11,12},
    xticklabels={}, 
    xmin=0, xmax=13,
    ymin=0, ymax=65,
]
\addplot[color=red!40, fill=red!40] coordinates {(3.1, 39.1) (4.1, 20.0) (5.1, 14.6) (6.1, 14.4) (7.1, 13.3) (8.1, 11.9) (9.1, 10.7) (10.1, 10.1) (11.1, 8.7) (12.1, 6.3)};
\end{axis}

\begin{axis}[
    axis y line=none,
    axis x line=top,
    width=\textwidth,
    height=.208\textheight,
    xlabel={\footnotesize \textbf{Layer Index}},
    xlabel style={at={(axis description cs:0.5,0.9)},anchor=south,font=\footnotesize},
    xtick=\empty,
    ytick=\empty,
    xmin=0.0, xmax=13,
    ymin=0.0, ymax=2,
]
\end{axis}

\begin{axis}[
    axis y line*=right,
    axis x line=top,
    width=\textwidth,
    height=.208\textheight,
    bar width=8pt,
    ybar = .05cm,
    ylabel = {\footnotesize \tikz \fill[red] (0,0) rectangle (0.02,0.1); \textbf{mIoU}},
    y tick label style={font=\tiny},
    x tick label style={font=\scriptsize},
    y label style={at={(axis description cs:1.14,.5)},anchor=south},
    xtick={1,2,3,4,5,6,7,8,9,10,11,12},
    x tick label style={font=\scriptsize, yshift=-0.60cm},
    xticklabels={\textbf{Layer3}, \textbf{Layer0}, \textbf{Layer8}, \textbf{Layer11}, \textbf{Layer4}, \textbf{Layer10}, \textbf{Layer7}, \textbf{Layer9}, \textbf{Layer2}, \textbf{Layer1}, \textbf{Layer6}, \textbf{Layer5}},
    xmin=0, xmax=13,
    ymin=0, ymax=65,
]
\end{axis}

\begin{axis}[
    axis y line=none,
    axis x line=none,
    width=\textwidth,
    height=.208\textheight,
    major x tick style = transparent,
    major y tick style = transparent,
    bar width=8pt,
    ybar = .05cm,
    ylabel = {\footnotesize mIoU},
    y tick label style={font=\tiny},
    y label style={at={(axis description cs:1.13,.5)},anchor=south},
    xmin=0, xmax=13,
    ymin=0, ymax=65,
]
\addplot[color=red, fill=red] coordinates {(1.1, 55.6) (2.1, 53.5)};
\end{axis}

\begin{axis}[
    axis y line*=right,
    width=\textwidth,
    height=.208\textheight,
    bar width=8pt,
    ybar = .05cm,
    ylabel = {\footnotesize \tikz \fill[red] (0,0) rectangle (0.02,0.1); \textbf{mIoU}},
    y tick label style={font=\tiny},
    x tick label style={font=\scriptsize},
    y label style={at={(axis description cs:1.14,.5)},anchor=south},
    xtick={0,1,2,3,4,5,6,7,8,9,10,11,12},
    x tick label style={font=\small, yshift=-0.05cm},
    xticklabels={\hspace{-10pt}\textbf{mIoU Rank}, (1st), (2nd), (3rd), (4th), (5th), (6th),(7th), (8th), (9th), (10th), (11th), (12th)},
    xmin=0, xmax=13,
    ymin=0, ymax=65,
]

\end{axis}

\begin{axis}[
    axis y line*=right,
    width=\textwidth,
    height=.208\textheight,
    bar width=8pt,
    ybar = .05cm,
    ylabel = {\footnotesize \tikz \fill[red] (0,0) rectangle (0.02,0.1); \textbf{mIoU}},
    y tick label style={font=\tiny},
    x tick label style={font=\scriptsize},
    y label style={at={(axis description cs:1.14,.5)},anchor=south},
    xtick={0,1,2,3,4,5,6,7,8,9,10,11,12},
    x tick label style={font=\small, yshift=-0.50cm},
    xticklabels={\hspace{-10pt}\textbf{InfoS Rank}, \textbf{(1st)}, \textbf{(2nd)}, \textbf{(5th)}, \textbf{(4th)}, \textbf{(7th)}, (12th), \textbf{(8th)}, \textbf{(9th)}, (3rd) , \textbf{(10th)} , (6th) , \textbf{(11th)}},
    xmin=0, xmax=13,
    ymin=0, ymax=65,
]

\end{axis}
\end{tikzpicture}

%% file: sec/4_results.tex
\section{Experimental Results} 
\subsection{Experimental Setup}
\noindent \textbf{Datasets.} 
We evaluate our method on four commonly used segmentation datasets: PASCAL-21~\cite{pascal}, COCO-Obj~\cite{ms_coco}, COCO-Stuffs-171~\cite{coco_stuff} and ADE-20K~\cite{zhou2017scene}. PASCAL-21 and COCO-Obj include a background category, while COCO-Stuffs-171 and ADE-20K do not, but are the most challenging due to the inclusion of stuffs classes. Following previous works~\cite{wang2025sclip, hajimiri2024pay}, the background class is represented as a list of possible background categories. Details about the class names used for single and multi-prompt setting are provided in the supplementary. For one-shot evaluation, we conduct five separate runs and average their mIoU. 

%

\vspace{0.05in}
\noindent \textbf{Implementation Details.} Implementation details regarding the model weights used for BLIP~\cite{li2022blip}, LLaVA~\cite{liu2024improved} and ALBEF~\cite{li2021align} are provided in the supplementary. We report results on images resized to a maximum side length of 512 for BLIP and ALBEF and 224 for LLaVA. 
By default, we use top-2 layers for all datasets and models except for ADE-20K where top-1 performs slightly better.
For ConvCRF, we use a kernel size of $15 \times 15$ and perform 20 iterations per image. We follow~\cite{teichmann2018convolutional} for other hyper-parameters pf ConvCRF.  Further details regarding fine-tuning is provided in the supplementary.


\begin{table}[t]
\centering
\resizebox{0.49\textwidth}{!}{
\begin{tabular}{lccc}
\toprule
\textbf{Layer Selection} & \textbf{PASCAL-21} & \textbf{COCO-Obj} & \textbf{COCO-171} \\
\midrule
\multicolumn{4}{l}{\emph{Training Free}} \\
\midrule
All layers (12 layers) & 42.8 & 35.9 & 25.1 \\
Random (1 layer) &  21.5 & 15.6 & 11.0\\
Naive (First 2 layers) & 48.2 & 37.2 & 25.5 \\
Naive (Last 2 layers) & 19.6 & 10.0 & 6.8 \\
InfoScore (Top-1) &  55.6 & 39.0 & 25.9 \\
InfoScore (Top-2) & \textbf{58.0} &\textbf{ 42.6} & \textbf{28.3} \\
InfoScore (Top-3) & 51.3 & 38.0 & 27.4 \\
InfoScore (Top-6) & 49.2 & 37.0 & 25.2 \\
\midrule
\multicolumn{4}{l}{\emph{One-shot Supervision}} \\
\midrule
InfoScore (Top-1) & 66.8 & 44.3 & 26.6 \\
InfoScore (Top-2) & \textbf{67.5} & \textbf{45.4} & \textbf{28.9} \\
\bottomrule
\end{tabular}
}
\vspace{-0.1in}
\caption{{\bf Ablation study on InfoScore-based layer selection.} Results show that using ensemble of attention maps from top-1, top-2, or top-3 layers ranked by InfoScore metric outperforms ensemble of all layers, random, or naive selection. Results are for single-prompt per class; best in \textbf{bold}.}
\label{tab:layer_selection_ablation}
\end{table}

\subsection{Comparison to the State of the Art} 

We compare against the state-of-the-art methods in open vocabulary segmentation. We mainly compare against training-free methods~\cite{luo2024emergent,wang2025sclip,hajimiri2024pay} and the ones that further fine-tune pre-trained VLMs on large-scale image-text pairs or pseudo annotations through weak supervision~\cite{xu2022groupvit,cha2023learning}. The choice for open vocabulary segmentation setup is motivated for the sake of fair comparison, since previous training-free methods had access to aligned vision-language data similar to our approach in its training-free mode. In Table~\ref{tab:comparison_ev}, it is shown that our training-free open-vocabulary method outperforms the state-of-the-art ProxyCLIP~\cite{lan2025proxyclip} on three challenging benchmark datasets COCO-Obj, COCO-171 and ADE-20K by 5.3\%, 1.6\% and 0.7\% respectively while achieving competitive performance on PASCAL-21.
Additionally, when provided with a single visual example and only a few iterations of fine-tuning on a small number of parameters, the segmentation perfromance significantly improves across all four datasets, particularly PASCAL-21, highlighting the value of visual aids, even with a single example, to clarify ambiguities that may arise from using textual prompts alone.  Examples include the stuffs classes in COCO-171 that suffer from being ambiguous, (e.g., \texttt{solid} or \texttt{structural}).

\begin{table}[t]
\centering

\setlength{\tabcolsep}{12pt}
\resizebox{0.45\textwidth}{!}{ 
\begin{tabular}{lccc}
\toprule
\textbf{Method} & \textbf{Layers} & \textbf{PASCAL-21} & \textbf{COCO-Obj} \\
\midrule
\multicolumn{4}{l}{\emph{Training Free}} \\
\midrule
BLIP & Top-1 & 55.6 & 39.0 \\
BLIP & Top-2 & \textbf{58.0} & \textbf{42.6} \\
ALBEF & Top-1 & 37.9 & 28.7 \\
ALBEF & Top-2 & 43.2 & 31.7 \\
LLaVA-1.5-7B & Top-1 & 41.8 & 24.7 \\
LLaVA-1.5-7B  & Top-2 & 40.7 & 24.5 \\
\midrule
\multicolumn{4}{l}{\emph{One-shot Supervision}} \\
\midrule
BLIP & Top-1 & 66.8 & 44.3 \\
BLIP & Top-2 & \textbf{67.5} & \textbf{45.4} \\
ALBEF & Top-1 & 65.1 & 37.7 \\
ALBEF & Top-2 & 65.5 & 38.7 \\
LLaVA-1.5-7B & Top-1 & 59.8 & 37.3 \\
LLaVA-1.5-7B  & Top-2 & 59.2 & 36.9 \\
\bottomrule
\end{tabular}
}
\vspace{-0.1in}
\caption{{\bf Ablation study comparing segmentation performance across different vision-language models (VLMs).} Fine-tuning with a single example significantly improves performance for BLIP~\cite{li2022blip}, ALBEF~\cite{li2021align} and LLaVA-1.5-7B~\cite{liu2024improved}. Results shown are for the single-prompt per class setting. Best results are \textbf{bolded}.}
\label{tab:vlms_ablation}
\vspace{-0.1in}
\end{table}

\subsection{Ablation Study}
\label{sec:ablation}
In this section, we focus on five ablations, followed by visual analysis. Ablations are conducted on the single prompt setting, except for multiple \textit{vs.} single ablation.

\vspace{0.05in}
\noindent
\textbf{InfoScore ranking and selection.} We ablate our novel InfoScore metric, which ranks and selects image-text attention layers, eliminating the need for hyperparameter tuning of layer choice. Table~\ref{tab:layer_selection_ablation} compares the ensemble of attention maps from all the layers in the VLM or randomly selected layers (Random) against our layer selection approach on BLIP. The results demonstrate that using an ensemble of the top attention maps, ranked by the proposed InfoScore, significantly outperforms the simple (All layers) and (Random) baselines across all datasets and settings.
Additionally, it shows that using the top-2 layers performs better than the top-1; performance begins to saturate at the top-3 and then declines with more layers ({\em e.g.}, top-6). Furthermore, our ranking outperforms naive baselines that select the first or last two layers. 




\begin{table}[t]
\centering
\resizebox{0.43\textwidth}{!}{
\begin{tabular}{lccc}
\toprule
\textbf{Method} & \textbf{Class VLM Score} & \textbf{PASCAL-21} & \textbf{COCO-Obj} \\
\midrule
\multicolumn{4}{l}{\emph{Training Free}} \\
\midrule
BLIP  &  \text{\sffamily $\times$} & 25.1 & 26.0 \\
BLIP  &  \checkmark & \textbf{58.0} &\textbf{42.6}\\
LLaVA-1.5-7B   &  \text{\sffamily $\times$} & 23.3 & 10.8 \\
LLaVA-1.5-7B   &  \checkmark & 40.7 & 24.5 \\

\midrule
\multicolumn{4}{l}{\emph{One-shot Supervision}} \\
\midrule
BLIP  &  \text{\sffamily $\times$} & 62.4 & 38.7 \\
BLIP  &  \checkmark & \textbf{67.5} & \textbf{45.4}\\
LLaVA-1.5-7B   &  \text{\sffamily $\times$} & 53.8 & 32.3 \\
LLaVA-1.5-7B   &  \checkmark & 59.2 & 36.9\\

\bottomrule
\end{tabular}
}
\vspace{-0.1in}
\caption{{\bf Ablation on the importance of image-text scoring.} image-text scoring relying on the class VLM scores for filtering significantly enhances performance in the training-free scenario and further improves results in the one-shot setting. As can be seen, it becomes less important with 1-shot fine-tuning. Results shown are in the single-prompt per class setting. Best are \textbf{bolded}.}
\label{tab:itm_selection_ablation}
\vspace{-0.1in}
\end{table}

\begin{table}[t]
\centering
\setlength{\tabcolsep}{14pt}
\resizebox{0.43\textwidth}{!}{
\begin{tabular}{lccc}
\toprule
\textbf{Method} & \textbf{PASCAL-21} & \textbf{COCO-Obj} & \textbf{COCO-171} \\
\midrule
\multicolumn{4}{l}{\emph{Training Free}} \\
\midrule
Single & 58.0 & 42.6 & \textbf{28.3} \\
Multiple & \textbf{60.2} & \textbf{42.8} & 28.1\\
\midrule
\multicolumn{4}{l}{\emph{One-shot Supervision}} \\
\midrule
Single & 67.5 & \textbf{45.4} & 28.9 \\
Multiple & \textbf{70.1} & 45.3& \textbf{29.3} \\
\bottomrule
\end{tabular}
}
\caption{\textbf{Ablation of single vs. multiple prompts with related words.} While multiple prompts improve segmentation accuracy on PASCAL-21, their impact is less significant on more challenging datasets, COCO-Obj and COCO-171. Best results are in \textbf{bold}.}
\label{tab:prompts_ablation}
\vspace{-1em}
\end{table}

\vspace{0.05in}
\noindent
\textbf{Extensibility across VLMs.} Table~\ref{tab:vlms_ablation} confirms the extensibility of our method and its ease of use across three VLMs including the recent ones relying on LLMs. It consistently shows that the use of a single visual example leads to considerable improvements across the three VLMs despite not introducing any new parameters or classifier. BLIP outperforms both ALBEF and LLaVA 1.5 on the two benchmarks.

\vspace{0.05in}
\noindent
\textbf{Image-text scoring.} Table~\ref{tab:itm_selection_ablation} ablates the impact of re-weighting the image-text attention maps using the class VLM scores already computed in the VLMs. The straightforward use of this re-weighting in our design results in significant improvements in segmentation performance for both training-free and 1-shot settings. The impact of class VLM scores becomes less important with one-shot tuning.

\noindent
\textbf{Multiple prompts.} We also ablate the use of a single prompt corresponding to each class name against the use of multiple prompts of related words to that specific class name in Table~\ref{tab:prompts_ablation}. It shows that multiple prompts either provide better or on-par performance to the single prompt across both training free and one-shot supervision modes. 

\vspace{0.05in}
\noindent
\textbf{Beyond one-shot performance.} Our method is scalable beyond one-shot. Please check supplementary for details.

\vspace{0.05in}
\noindent
\textbf{One-shot fine-tuning parameter selection.} Finally, we ablate which parameters to fine-tune during our one-shot adaptation mechanism. Table~\ref{tab:parameter_selection_effect} compares five variants with BLIP; (i) the use of all parameters in the VLM, (ii) fine-tuning all the parameters of the multi-modal encoder, (iii) only fine-tuning the word embeddings, (iv) only fine-tuning the top-2 selected cross-attention layers and (v) fine-tuning both the word embeddings and top-2 cross-attention layers. It shows that the use of both the word embeddings and top-2 layers outperforms the other variants.

\begin{table}[t]
\centering
\setlength{\tabcolsep}{14pt}
\resizebox{0.9\columnwidth}{!}{%
\begin{tabular}{lcc}
\toprule
\textbf{Params Optimized} & \textbf{PASCAL-21} & \textbf{COCO-Obj} \\
\midrule
All params & 60.9 & 41.9 \\
Multi-modal encoder params & 64.1 & 40.6 \\
Word embed only & 63.2 & 43.4 \\
Top-2 CA layers only & 65.0 & 44.4 \\
Word embed + top-2 CA layers & \textbf{67.5} & \textbf{45.4} \\
\bottomrule
\end{tabular}%
}
\caption{{\bf Parameter selection in few-shot fine-tuning.} Shows the effectiveness of selectively fine-tuning word embeddings and the parameters of top-2 layers ranked by InfoScore. Results shown are in the single-prompt per class setting. Best results are \textbf{bolded}.}
\label{tab:parameter_selection_effect}
\vspace{-1em}
\end{table}

\begin{figure}[t!]
\centering
    \includegraphics[width=0.5\textwidth]{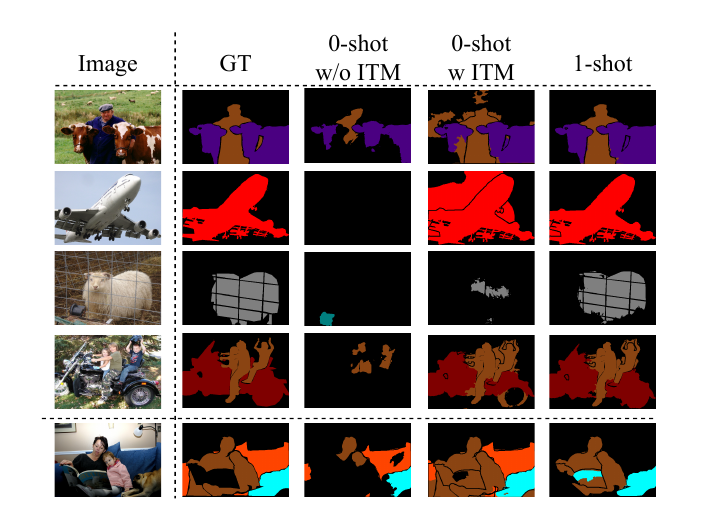}
    \vspace{-2.5em}
    \caption{{\bf Qualitative Results on PASCAL-21:} Shown are results from zero-shot model w/o image-text scoring (3rd column), zero-shot model w/ image-text scoring (4th column), and one-shot fine-tuning (5th column). The final row shows an example where the zero-shot prediction outperformed the fine-tuned one-shot model. For all variants, we ensemble the top-2 layers ranked by InfoScore.}
    \label{fig:qualitative}
    \vspace{-1em}
\end{figure}

\vspace{0.05in}
\noindent
\textbf{Visual analysis.} Figure~\ref{fig:qualitative} presents the qualitative results of our approach across different settings, illustrating the impact of image-text scoring and one-shot fine-tuning. As shown, the model without image-text scoring consistently under-segments across all images, labeling most of the pixels as background. Applying image-text scoring for filtering notably improves performance across all images, as it helps the model better focus on relevant areas. 
One-shot fine-tuning further enhances segmentation accuracy, consistently producing more precise segmentations. However, the last row highlights a failure case for the one-shot fine-tuning, where potential bias from the single example of the class \texttt{couch} led to mis-classification. More qualitative results are provided in the Suppl. 

%% file: sec/5_conc.tex
\section{Conclusion}
Our work extends state-of-the-art approaches that perform segmentation by using vision foundation models by eliminating the need for abundant segmentation labels. We leverage the strength of a single visual example to better disambiguate categories beyond their textual names. This, combined with our proposed InfoScore metric, reduces reliance on intensive prompt engineering or tuning of the layers/heads selected for segmentation in VLMs. Our approach can operate in both training-free and one-shot fine-tuning settings, with the latter achieving significant gains on four benchmarks and demonstrating compatibility across three different VLMs.

%% file: sec/X_suppl.tex
\clearpage
\setcounter{page}{1}
\maketitlesupplementary
\begin{abstract}
This document provides supplementary material to support our main work. Section~\ref{sec:impl} describes additional implementation details. Section~\ref{sec:infoscore} provides additional analysis on our proposed InfoScore metric for the layer ranking and selection. Section~\ref{sec:fewshot} demonstrates the scalability of our approach across multiple shots, extending beyond a single visual example. Section~\ref{sec:param_efficiency} discusses the number and proportion of parameters optimized relative to the total parameter count. Section~\ref{sec:postproc} presents an ablation study on the post-processing methods used. Section~\ref{sec:partial_vocab} discusses how our approach can be applied to unseen classes when we provide a 1-shot example for only 25\% of the classes in the vocabulary. Section~\ref{sec:qual} includes additional qualitative results on the COCO-Obj dataset. Finally, Section~\ref{sec:prompts} provides further details on the multiple-prompt settings employed throughout the study.
\end{abstract}


\section{Implementation Details} 
\label{sec:impl}
For BLIP~\cite{li2022blip} and LLaVA~\cite{liu2024improved}, we rely on the \texttt{\small HuggingFace}\footnote{\url{https://huggingface.co/docs/transformers/en/model_doc/blip}} library. For ALBEF~\cite{li2021align}, we use \texttt{\small LAVIS}\footnote{\url{https://github.com/salesforce/LAVIS}}. 

\vspace{0.05in}
\noindent \textbf{BLIP.} For BLIP we use pretrained model weights of \texttt{Salesforce/blip-itm-large-coco}. It uses ViT-L/16 backbone pre-trained on ImageNet as vision encoder and BERT~\cite{devlin2019bert} as image-grounded text encoder. The cross-attention layers of BERT take image embeddings as key, value. There are 12 such cross-attention layers and the number of heads is 12. The BLIP model is trained for image-text understanding using a combination of Image-Text Matching (ITM) and Image-Text Contrastive (ITC) Loss on 129M image-text pairs  The ITM head of BLIP model is used for class-scoring. 

\vspace{0.05in}
\noindent \textbf{ALBEF.} For ALBEF we use \texttt{albef-retrieval-coco} checkpoint as pre-trained model weights. It uses ViT-B/16 pretrained on ImageNet. Similar to BLIP they are also trained with ITM and ITC Loss but on a much smaller number of image-text pairs of 14.1M. It has six multi-modal layers where there is explicit cross-attention between image and text. 

\vspace{0.05in}
\noindent \textbf{LLaVA.} For LLaVA we use \texttt{llava-1.5-7b-hf} as pre-trained model weights. It uses CLIP VIT-L/14 as a vision encoder which is pre-trained with 400M image-text pairs. The image embeddings are first passed into a projection layer which are then passed to a large language model relying on Vicuna-7B.

\vspace{0.05in}
\noindent \textbf{One-shot fine-tuning details} For one-shot supervision we use a batch size of two for COCO-Obj, PASCAL-21 and ADE-20K and a batch size of six for COCO-171. For COCO-Obj, PASCAL-21 and ADE-20K we use training images cropped to maximum width of 512, while for COCO-171 we use a resolution of 256 for training but 512 during inference time. We fine-tune all models (BLIP, ALBEF, LLaVA) on a single A40 GPU with 48GB memory. In order to fine-tune LLaVA we use Low-rank adapter (LoRA) and 4-bit quantization. We use a learning rate of $2 \times 10^{-4}$ for PASCAL-21 and COCO-Obj, and $5 \times 10^{-5}$ for COCO-171 and ADE-20K. We train COCO-Obj for two epochs, COCO-171 and ADE-20K for three epochs and PASCAL-21 for five epochs.

\vspace{0.05in}
\noindent \textbf{Additional details.} For the random baseline we report in our ablations, we assume a uniform sampling strategy for selecting each cross-attention layer individually, evaluate its final mIoU, and then report the average.


\section{InfoScore Additional Analysis}
\label{sec:infoscore}

In this section, we provide additional analysis on our proposed InfoScore metric. In Table~\ref{tab:layer_ranking_table3} we show our InfoScore ranking for ALBEF compared to the mIoU ranking similar to the main submission Figure 3 in a tabular format. The scores are evaluated across PASCAL-21 validation set. It clearly shows that our unsupervised ranking perfectly aligns with the mIoU ranking across all the layers, without requiring any annotations. In Table~\ref{tab:albef_layer_selection_ablation}, we demonstrate the effectiveness of using InfoScore for layer selection on ALBEF. Top-1, top-2 and top-3 all outperform the random layer selection, averaging all layers and the naive selection of the last two layers. Ensembling top-2 or top-3 layers are better than the naive selection of the first two layers.

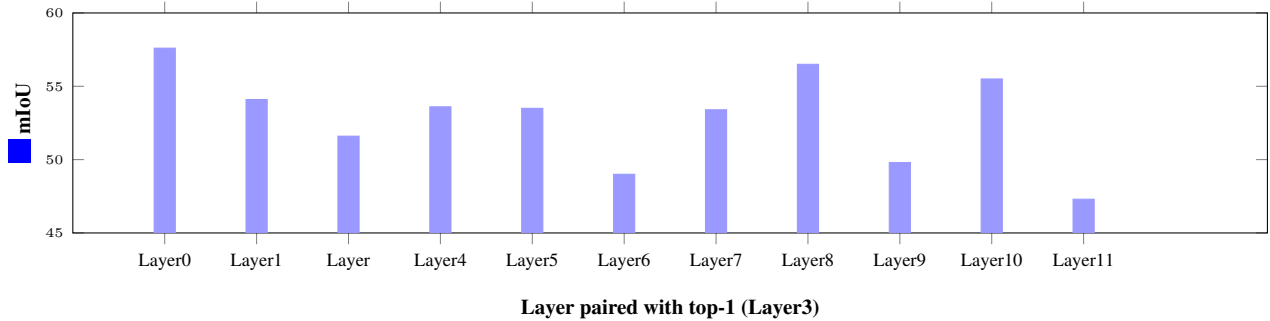
\begin{figure*}[ht]
\centering
\input{Images/plot_infoscore_supplablation}
\caption{{\bf Ablation on second pair.} Ablation study on the mIoU pairing the Top-1 layer (Layer3) with all other layers using BLIP and evaluated on PASCAL-21.}
\label{fig:ablatioin_infoscore_paired_layer}
\end{figure*}

Furthermore, we ablate a challenging setting where we use the top-1 layer of BLIP identified with our InfoScore metric paired with all the other layers in BLIP evaluated on PASCAL-21 dataset. Figure~\ref{fig:ablatioin_infoscore_paired_layer} shows that pairing the top-1 layer (Layer3) with the second best layer (Layer0), outperforms all the other pairs. Thus, it confirms the benefits from our proposed approach even in the challenging setting.

In Table~\ref{tab:llava_layer_selection_ablation}, we demonstrate the effectiveness of selecting layers for an ensemble using our InfoScore with LLaVA 1.5. The top-$k$ layers consistently outperform randomly selecting one layer or naively choosing the first or last two layers. Empirically, the top-6 and top-16 layers perform the best. The top-16 layer selected by InfoScore outperforms naively selecting the first and last 16 layers, further highlighting the ranking effectiveness of InfoScore. Although an ensemble of all 32 layers proves to be reasonably effective, the top-6 and top-16 ensembles outperform it. 



\section{Few-shot Ablation}
\label{sec:fewshot}
In this section, we examine the scalability of our method beyond the one-shot setting. Table~\ref{tab:shots-ablation} compares the 3-shot and 5-shot performance against the 1-shot performance for both single and multi-prompt settings on the PASCAL-21 and COCO-Obj datasets. The reported results are averaged over five runs. As expected, increasing the number of shots leads to improvements of approximately 1-2\% on both datasets. Additionally, the performance gap between single and multi-prompt settings diminishes at higher shot counts, indicating that our method does not heavily rely on prompt engineering, especially when incorporating a few visual demonstrations beyond a single example.

\begin{table}[t]
\centering
\setlength{\tabcolsep}{12pt}
\resizebox{0.9\columnwidth}{!}{
\begin{tabular}{cccc}
\toprule
\textbf{Layer} & \textbf{InfoScore Rank} & \textbf{mIoU Rank} & \textbf{mIoU} \\ 
\midrule
L1 & 3rd & 3rd & 23.3 \\
L2 & 1st & 1st & 37.9 \\
L3 & 2nd & 2nd & 33.3 \\
L4 & 4th & 4th & 11.5 \\
L5 & 6th & 6th & 6.7 \\
L6 & 5th & 5th & 10.7 \\
\bottomrule
\end{tabular}
}
\caption{The ranking of ALBEF layers, evaluated using the proposed InfoScore metric, which does not rely on ground-truth annotation from the PASCAL-21 validation set (1449 images). We compare this with the mIoU ranking sorting the actual mIoU in descending order. Our ranking aligns perfectly with the mIoU ranking across all the layers.}
\label{tab:layer_ranking_table3}
\end{table}

\begin{table}
\resizebox{0.45\textwidth}{!}{ 
\begin{tabular}{p{8cm}c} 
\toprule
\textbf{Layer Selection} & \textbf{mIoU}  \\
\midrule
All layers (6 layers) & 34.3  \\
Random (1 layer) & 20.6 \\
Naive (First 2 layers) & 38.4\\
Naive (Last 2 layers) & 11.0 \\
InfoScore (Top-1) &  37.9\\
InfoScore (Top-2) & \textbf{43.2}\\
InfoScore (Top-3) & 39.6 \\

\bottomrule
\end{tabular}
}
\caption{\textbf{Ablation on InfoScore-based automatic layer selection of ALBEF in training free setting on PASCAL-21.} ALBEF has a total of six multi-modal layers. Selecting the attention maps from the top-2 or top-3 layers is better random selection, naive selection or taking an average of all the layers. Best results are {\bf bolded} .}
\label{tab:albef_layer_selection_ablation}
\end{table}

\begin{table}[t]
\centering
\tiny

\resizebox{0.5\textwidth}{!}{ 
\begin{tabular}{p{4cm}c} 
\toprule
\textbf{Layer Selection} & \textbf{mIoU}  \\
\midrule
All layers (32 layers) & 43.6  \\
Random (1 layer) & 31.4 \\
Naive (First 2 layers) & 20.0\\
Naive (Last 2 layers) & 29.0 \\
Naive (First 16 layers) & 40.9\\
Naive (Last 16 layers) & 42.1 \\
\textcolor{black}{InfoScore (Top-1)} &  41.8\\
\textcolor{black}{InfoScore (Top-2)} & 40.7 \\
\textcolor{black}{InfoScore (Top-3)} & 42.2 \\
\textcolor{black}{InfoScore (Top-6)} & \textbf{43.9}\\
\textcolor{black}{InfoScore (Top-16)}  & \textbf{44.0} \\

\bottomrule
\end{tabular}
}
\caption{\textbf{Ablation on InfoScore-based automatic layer selection of LLaVA-1.5-7B in a training free setting on PASCAL-21.} LLaVA-1.5-7B has 32 layers. Consistent to other VLMs, InfoScore based top-$K$ selection, outperforms random 1 layer or naive $K$-layer selection for same number of $K$. Emperically, top-6 and top-16 layers perform the best and are better than taking average of attention maps of all layers. Best results are in \textbf{bold}.}
\label{tab:llava_layer_selection_ablation}
\end{table}

\begin{table}[t]
\centering
\small
\setlength{\tabcolsep}{20pt}
\resizebox{0.9\columnwidth}{!}{
\begin{tabular}{lcc}
\toprule
\textbf{Method} & \textbf{PASCAL-21} & \textbf{COCO-Obj}  \\
\midrule
\multicolumn{3}{l}{\emph{1-shot}  (Ave. Across 5 runs)} \\
\midrule
Single & 67.5 & 45.4 \\
Multiple & 70.1 & 45.3 \\
\midrule
\multicolumn{3}{l}{\emph{3-shot}  (Ave. Across 5 runs)} \\
\midrule
Single & 72.2 & 46.7  \\
Multiple & 72.1 & 46.3 \\
\midrule
\multicolumn{3}{l}{\emph{5-shot}  (Ave. Across 5 runs)} \\
\midrule
Single & \underline{73.0} & \underline{47.2}  \\
Multiple & \textbf{73.2} & \textbf{47.3} \\
\bottomrule
\end{tabular}
}
\caption{\textbf{Ablation with increased number of shots for single and multi-prompt setting.} Our model scales as the number of shots increases from 1 to 5 on both COCO-Obj and PASCAL-21 datasets. The impact of multiple prompts is further reduced with increasing the number of shots. Best results are in \textbf{bold}, and second-best results are \underline{underlined}.}
\label{tab:shots-ablation}
\end{table}

\begin{table}[t]
\centering
\small
\resizebox{\columnwidth}{!}{
\begin{tabular}{lcccc}
\toprule
\textbf{PostProc} & \textbf{Inference Time}  &\textbf{PASCAL-21} & \textbf{COCO-Obj}  \\

\midrule
ConvCRF~\cite{teichmann2018convolutional} & \textbf{0.4s} & \textbf{60.2} & \textbf{42.8} \\
PAMR~\cite{araslanov2020single} & 3.6s  & 59.6 & 42.2 \\

\bottomrule
\end{tabular}
}
\caption{\textbf{Ablation with two different post-processing method in the training-free setting.}  ConvCRF~\cite{teichmann2018convolutional} marginally outperforms PAMR~\cite{araslanov2020single} while being significantly faster at inference. Even with PAMR post-processing, our training-free setting outperforms state-of-the-art training-free approaches like NACLIP~\cite{hajimiri2024pay} on COCO-Obj by nearly 6\%. Best results are in \textbf{bold}.}
\label{tab:postproc_ablation}
\end{table}

\begin{table}[t]
\centering
\resizebox{\columnwidth}{!}{
\begin{tabular}{lcccccc}
\toprule
 & \multicolumn{5}{c}{\textbf{Unseen mIoU}} \\ 
\cmidrule(lr){2-6}
\textbf{Percent of Seen Classes} & \textbf{S1} & \textbf{S2} & \textbf{S3} & \textbf{S4} & \textbf{Ave} \\
\midrule
0\% (zero-shot) & 63.6 & 57.2 & \textbf{59.9} & 59.5 & 60.1 \\
25\% (one-shot) & \textbf{64.8} & \textbf{60.0} & 59.0 & \textbf{62.8} & \textbf{61.7} \\
\bottomrule
\end{tabular}
}
\caption{\textbf{Ablation study comparing fine-tuning with 1-shot learning on 25\% of the classes (75\% unseen) against zero-shot performance on the unseen classes in PASCAL-21.} The seen and unseen classes are divided into four non-overlapping splits. Fine-tuning with just one visual example per class for 25\% of the classes improves performance on the 75\% unseen categories.}
\label{tab:25_percent_transfer}
\end{table}

\section{Selective Parameter Optimization}
\label{sec:param_efficiency}
BLIP consists of approximately 446.13 million parameters. As mentioned in the main paper, we selectively fine-tune only the top-$K$ cross-attention layers and the word embeddings corresponding to the prompts. While BLIP has a vocabulary size of 30,524 words, in practice, we fine-tune only the embeddings of the words appearing in the prompts—up to approximately 250 word embeddings. Overall, this selective approach results in the fine-tuning of approximately 5.7 million parameters out of 446.13 million, representing just 1.28\% of the total parameters. Moreover, we do not introduce any additional parameters in either the training-free or few-shot fine-tuning settings. Our approach leverages the existing VLM architecture to generate segmentation maps without adding any new parameters, demonstrating its efficiency and scalability.

\section{Post-processing Ablation}
\label{sec:postproc}
In this section, we compare post-processing with ConvCRF~\cite{teichmann2018convolutional} and PAMR~\cite{araslanov2020single}, used by recent methods~\cite{wang2025sclip,hajimiri2024pay}. Table~\ref{tab:postproc_ablation} shows that our approach is robust to the choice of post-processing, with ConvCRF providing a slight improvement and being nine times faster than PAMR. In the zero-shot setting on COCO-Obj, our method outperforms the second-best approach, NACLIP~\cite{hajimiri2024pay}, by 6\%, regardless of post-processing. These results confirm that our approach consistently outperforms state-of-the-art methods, independent of the post-processing strategy.

\section{One-shot Tuning with Partial Vocabulary}
\label{sec:partial_vocab}
Table~\ref{tab:25_percent_transfer} shows the impact of 1-shot fine-tuning for only a small portion of total vocabulary and how the performance is translated to unseen categories. Specifically, we fine-tune using only single visual example on 25\% of the total classes, leaving 75\% of the classes entirely unseen during fine-tuning. The seen and usneen classes are divided into four non-overlapping splits, where for each split we fine-tune with 1-shot data for 25\% of the categories and evaluate for the rest of the 75\%. 

The results demonstrate that even with such minimal supervision—just one visual example per class for 25\% of the categories—the fine-tuned model significantly outperforms the zero-shot baseline on the remaining 75\% unseen categories. This improvement highlights the model’s ability to generalize knowledge learned from a small subset of classes to the broader unseen vocabulary, benefiting from the shared semantic structure among the categories in the dataset.

This finding underscores the effectiveness of leveraging a limited amount of labeled data to enhance performance in scenarios where many classes remain unlabeled, offering a practical approach for improving results in resource-constrained settings.

\section{Additional Qualitative Results}
\label{sec:qual}

\begin{figure*}[t]
\centering
    \includegraphics[width=0.65\textwidth]{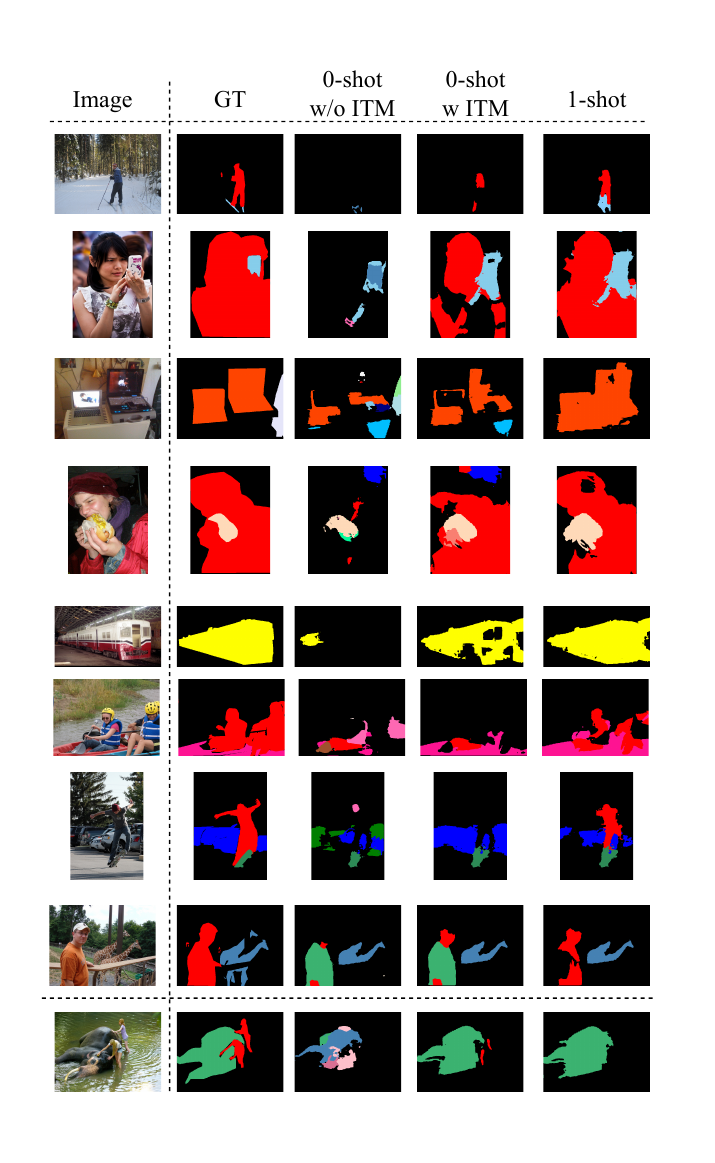}
    \vspace{-2.5em}
    \caption{{\bf Qualitative Results on COCO-Obj:} Shown are results from zero-shot model w/o image-text scoring for filtering (3rd column), zero-shot model w/ image-text scoring (4th column), and one-shot fine-tuning (5th column). The final row shows an example where the zero-shot prediction outperformed the fine-tuned one-shot model. For all variants, we ensemble the top-2 layers ranked by InfoScore.}
    \label{fig:qualitative_coco}
    \vspace{-1em}
\end{figure*}

In this section, we show additional results on COCO-Obj dataset in Fig.~\ref{fig:qualitative_coco}. It shows eight successful scenarios, where the variant without ITM re-weighting either under-segments or mis-classifies the objects in the scene. On the other hand, the one-shot variant with only a single visual example shows considerable improvement in segmenting the objects in the scene and overcomes the aforementioned issues. Looking specifically at the second and seventh rows we see that the one-shot variant resolved the confusion on whether to segment the clothes as part of class \texttt{Person} or not. Since by definition of the word itself it might exclude the clothes as another class, but with the class definition of COCO-Obj these are to be considered part of \texttt{Person} class.

\section{Multiple Prompts Details}
\label{sec:prompts}
In this section, we add the detailed prompts used in the multiple prompts setting. Our prompt takes the form \texttt{[Image of \{class\}.]}, where the \texttt{\{class\}} is the class name in the single prompt setting. For single- prompt per category setting, we conduct minor modifications on the names of certain categories. For instance, in COCO-Obj, we correct the misspelling of \texttt{\small hair drier} to \texttt{\small hair dryer}. In COCO-171, we remove ambiguous suffixes ({\em e.g.}, \texttt{\small -other, -stuff}) and rename classes like \texttt{\small floor-wood} to \texttt{\small wooden floor}. 

In the multiple prompts variant we rely on synonyms, hyponyms and/or plural for the class name as additional prompts. Note that the list of classes for the background is borrowed from the implementation of SCLIP~\cite{wang2025sclip}\footnote{\url{https://github.com/wangf3014/SCLIP/blob/main/configs/cls_coco_object.txt}}
and NACLIP~\cite{hajimiri2024pay}\footnote{\url{https://github.com/sinahmr/NACLIP/blob/main/configs/cls_coco_object.txt}}. Table~\ref{tab:pascal_prompts} shows the prompts used in PASCAL-21 in the multiple setting, while Table~\ref{table:coco_prompts} shows the prompts used in both COCO-Obj (80 classes) and COCO-171 (171 classes). The table shows both the things classes, which are common in both datasets, and the stuff classes, which only exist in COCO-171 dataset. 

\begin{table}[h]
\centering
\resizebox{0.45\textwidth}{!}{
\begin{tabular}{|lp{5cm}|}
\hline
Class Id and Prompt & Additional Prompts \\ \hline
\textbf{0:} \texttt{background$^*$} & \texttt{sky, wall, tree, wood, grass, road, sea, river, mountain, sands, desk, building, cloud, lamp, door, window, wardrobe, ceiling, shelf, curtain, stair, floor, hill, rail, fence} \\ 
\textbf{1:} \texttt{airplane} & \texttt{aeroplane,  jet, airplanes, plane, aeroplanes, jets, planes} \\
\textbf{2:} \texttt{bicycle} & \texttt{bicycles, bike, bikes} \\
\textbf{3:} \texttt{bird} & \texttt{birds} \\
\textbf{4:} \texttt{boat} & \texttt{boats, yacht, ship, ships,speedboat, speedboats,  yachts} \\
\textbf{5:} \texttt{bottle} & \texttt{bottles} \\
\textbf{6:} \texttt{bus} & \texttt{buses, coach, coaches} \\
\textbf{7:} \texttt{car} & \texttt{cars} \\
\textbf{8:} \texttt{cat} & \texttt{cats} \\
\textbf{9:} \texttt{chair} & \texttt{chairs, dining chair} \\
\textbf{10:} \texttt{cow} & \texttt{cows, cattle} \\
\textbf{11:} \texttt{dining table} & \texttt{dining tables} \\
\textbf{12:} \texttt{dog} & \texttt{dogs} \\
\textbf{13:} \texttt{horse} & \texttt{horses} \\
\textbf{14:} \texttt{motorcycle} & \texttt{motorcycles, motorbike, motorbikes} \\
\textbf{15:} \texttt{person} & \texttt{people, man, woman, men, women, boys, girls, child, children, boy, person in shirt, person in jeans, person in dress,person in sweater, person in skirt,person in jacket} \\
\textbf{16:} \texttt{potted plant} & \parbox{5cm}{\texttt{potted plants, indoor plants, house plants}} \\
\textbf{17:} \texttt{sheep} & - \\
\textbf{18:} \texttt{couch} & \texttt{sofa, couches, sofa} \\
\textbf{19:} \texttt{train} & \parbox{5cm}{\texttt{trains, railcar, railcars}} \\
\textbf{20:} \texttt{tv} & \texttt{television, television set, television monitor, tv monitor, monitor, television, television screen, TVs} \\
\hline
\end{tabular}}
\caption{\textbf{List of prompts for Multi-Prompt Settings for PASCAL-21.}
The symbol $^*$ for 'background' indicates that additional prompts are used to represent the background classes for the single prompt setting, following~\cite{wang2025sclip, hajimiri2024pay}. The list of classes for the background is borrowed from the implementation of SCLIP~\cite{wang2025sclip} and NACLIP~\cite{hajimiri2024pay}. Our prompt takes the form \texttt{[Image of \{class\}.]}, where the \texttt{\{class\}} is given above.}
\label{tab:pascal_prompts}
\end{table}

\onecolumn

\begin{longtable}{|c|p{3.7cm} p{10cm}|}
\caption{\textbf{List of prompts for Single and Multi-Prompt Settings for COCO-Obj (Things Classes) and COCO-171 (Things + Stuff Classes)}. The symbol $^*$ for 'background' indicates that additional prompts are used to represent the background classes for the single prompt setting, following ~\cite{wang2025sclip, hajimiri2024pay}. The list of classes for the background is borrowed from the implementation of SCLIP~\cite{wang2025sclip} and NACLIP~\cite{hajimiri2024pay}. Our prompt takes the form \texttt{[Image of \{class\}.]}, where the \texttt{\{class\}} is given below. It is to be noted that COCO-171 does not have any background class and the background prompts are used for COCO-Obj only.} 
\label{table:coco_prompts}
\\ \hline
 & \multicolumn{1}{|l}{Class Id and Prompt} & \multicolumn{1}{l|}{Additional Prompts (Multi-prompt)} \\ \hline
\endfirsthead
\hline
 & \multicolumn{1}{|l}{Class Id and Prompt} & \multicolumn{1}{l|}{Additional Prompts (Multi-prompt)} \\ \hline
\endhead
\hline
\endfoot
\hline
\endlastfoot
Background & \multicolumn{1}{p{3.7cm}}{\textbf{0:} \texttt{background$^*$}} & \parbox{10cm}{\texttt{sky, wall, tree, wood, grass, road, sea, river, mountain, sands, desk, building, cloud, lamp, door, window, wardrobe, ceiling, shelf, curtain, stair, floor, hill, rail, fence}} 
\\ \hline
\multirow{51}{*}{\rotatebox{90}{\centering Things Classes}} 
& \textbf{1:} \texttt{person} & \texttt{people, man, woman, child, children, boy, girl} \\ 
& \textbf{2:} \texttt{bicycle} & \texttt{bicycles, bike} \\ 
& \textbf{3:} \texttt{car} & \texttt{cars.} \\ 
& \textbf{4:} \texttt{motorcycle} & \texttt{motorcycles, motorbike} \\ 
& \textbf{5:} \texttt{airplane} & \texttt{airplanes, aeroplane, aircraft} \\ 
& \textbf{6:} \texttt{bus} & \texttt{buses, coach} \\ 
& \textbf{7:} \texttt{train} & - \\ 
& \textbf{8:} \texttt{truck} & \texttt{trucks, lorry} \\ 
& \textbf{9:} \texttt{boat} & \texttt{ship, boats, yacht, sailboat, speedboat} \\ 
& \textbf{10:} \texttt{traffic light} & - \\ 
& \textbf{11:} \texttt{fire hydrant} & - \\ 
& \textbf{12:} \texttt{stop sign} & - \\ 
& \textbf{13:} \texttt{parking meter} & - \\ 
& \textbf{14:} \texttt{bench} & \texttt{benches} \\ 
& \textbf{15:} \texttt{bird} & \texttt{birds} \\ 
& \textbf{16:} \texttt{cat} & \texttt{cats, kitten} \\ 
& \textbf{17:} \texttt{dog} & \texttt{dogs, puppy} \\ 
& \textbf{18:} \texttt{horse} & \texttt{horses} \\ 
& \textbf{19:} \texttt{sheep} & - \\ 
& \textbf{20:} \texttt{cow} & \texttt{cows, cattle} \\ 
& \textbf{21:} \texttt{elephant} & \texttt{elephants} \\ 
& \textbf{22:} \texttt{bear} & \texttt{bears} \\ 
& \textbf{23:} \texttt{zebra} & \texttt{zebras} \\ 
& \textbf{24:} \texttt{giraffe} & \texttt{giraffes} \\ 
& \textbf{25:} \texttt{backpack} & \texttt{backpacks} \\
& \textbf{26:} \texttt{umbrella} & \texttt{parasol, umbrellas} \\
& \textbf{27:} \texttt{handbag} & - \\
& \textbf{28:} \texttt{tie} & \texttt{necktie} \\ 
& \textbf{29:} \texttt{suitcase} & - \\ 
& \textbf{30:} \texttt{frisbee} & - \\ 
& \textbf{31:} \texttt{skis} & \texttt{ski} \\ 
& \textbf{32:} \texttt{snowboard} & \texttt{snowboards} \\ 
& \textbf{33:} \texttt{sports ball} & \texttt{ball, sports balls} \\ 
& \textbf{34:} \texttt{kite} & \texttt{kites} \\ 
& \textbf{35:} \texttt{baseball bat} & - \\ 
& \textbf{36:} \texttt{baseball glove} & - \\ 
& \textbf{37:} \texttt{skateboard} & \texttt{skateboards} \\ 
& \textbf{38:} \texttt{surfboard} & \texttt{surfboards} \\ 
& \textbf{39:} \texttt{tennis racket} & \texttt{racket, tennis rackets, racquet} \\ 
& \textbf{40:} \texttt{bottle} & \texttt{bottles} \\ 
& \textbf{41:} \texttt{wine glass} & - \\ 
& \textbf{42:} \texttt{cup} & \texttt{cups} \\ 
& \textbf{43:} \texttt{fork} & \texttt{forks} \\ 
& \textbf{44:} \texttt{knife} & \texttt{knives} \\ 
& \textbf{45:} \texttt{spoon} & - \\ 
\multirow{36}{*}{\rotatebox{90}{\centering Things Classes}} & \textbf{46:} \texttt{bowl} & \texttt{dish} \\ 
& \textbf{47:} \texttt{banana} & \texttt{bananas} \\ 
& \textbf{48:} \texttt{apple} & \texttt{apples} \\ 
& \textbf{49:} \texttt{sandwich} & \texttt{sandwiches} \\ 
& \textbf{50:} \texttt{orange} & \texttt{oranges} \\ 
& \textbf{51:} \texttt{broccoli} & - \\ 
& \textbf{52:} \texttt{carrot} & \texttt{carrots} \\ 
& \textbf{53:} \texttt{hotdog} & \texttt{hotdogs, sausage} \\ 
& \textbf{54:} \texttt{pizza} & - \\ 
& \textbf{55:} \texttt{donut} & \texttt{donuts} \\ 
& \textbf{56:} \texttt{cake} & \texttt{cakes} \\ 
& \textbf{57:} \texttt{chair} & \texttt{chairs} \\ 
& \textbf{58:} \texttt{couch} & \texttt{sofa} \\ 
& \textbf{59:} \texttt{potted plant} & \texttt{indoor plant} \\ 
& \textbf{60:} \texttt{bed} & - \\ 
& \textbf{61:} \texttt{dining table} & - \\ 
& \textbf{62:} \texttt{toilet} & - \\ 
& \textbf{63:} \texttt{tv} & \texttt{television, tvs, television screen} \\ 
& \textbf{64:} \texttt{laptop} & - \\ 
& \textbf{65:} \texttt{mouse} & \texttt{computer mouse} \\ 
& \textbf{66:} \texttt{remote} & \texttt{remote control, remotes} \\ 
& \textbf{67:} \texttt{keyboard} & - \\ 
& \textbf{68:} \texttt{cell phone} & \texttt{cell phones, mobile phone} \\ 
& \textbf{69:} \texttt{microwave} & - \\ 
& \textbf{70:} \texttt{oven} & - \\ 
& \textbf{71:} \texttt{toaster} & - \\ 
& \textbf{72:} \texttt{sink} & - \\ 
& \textbf{73:} \texttt{refrigerator} & \texttt{fridge} \\ 
& \textbf{74:} \texttt{book} & \texttt{books} \\ 
& \textbf{75:} \texttt{clock} & - \\ 
& \textbf{76:} \texttt{vase} & - \\ 
& \textbf{77:} \texttt{scissors} & - \\ 
& \textbf{78:} \texttt{teddy bear} & \texttt{teddy} \\ 
& \textbf{79:} \texttt{hair dryer} & \texttt{blow dryer} \\ 
& \textbf{80:} \texttt{toothbrush} & - \\ \hline

\multirow{16}{*}{\rotatebox{90}{\centering Stuffs Classes}}& \textbf{81:} \texttt{banner} & - \\ 
& \textbf{82:} \texttt{blanket} & - \\ 
& \textbf{83:} \texttt{branch} & \texttt{branches, tree branch} \\ 
& \textbf{84:} \texttt{bridge} & - \\ 
& \textbf{85:} \texttt{building} & \texttt{buildings} \\ 
& \textbf{86:} \texttt{bush} & \texttt{bushes} \\ 
& \textbf{87:} \texttt{cabinet} & \texttt{storage, wall cabinet} \\ 
& \textbf{88:} \texttt{cage} & - \\ 
& \textbf{89:} \texttt{cardboard} & - \\ 
& \textbf{90:} \texttt{carpet} & - \\ 
& \textbf{91:} \texttt{ceiling} & - \\ 
& \textbf{92:} \texttt{tile ceiling} & - \\ 
& \textbf{93:} \texttt{cloth} & - \\ 
& \textbf{94:} \texttt{clothes} & - \\ 
& \textbf{95:} \texttt{clouds} & - \\ 
& \textbf{96:} \texttt{counter} & - \\ 
\multirow{51}{*}{\rotatebox{90}{\centering Stuffs Classes}} & \textbf{97:} \texttt{cupboard} & - \\ 
& \textbf{98:} \texttt{curtain} & - \\ 
& \textbf{99:} \texttt{desk} & - \\ 
& \textbf{100:} \texttt{dirt} & - \\ 
& \textbf{101:} \texttt{door} & - \\ 
& \textbf{102:} \texttt{fence} & - \\ 
& \textbf{103:} \texttt{marble floor} & - \\ 
& \textbf{104:} \texttt{floor} & - \\ 
& \textbf{105:} \texttt{stone floor} & - \\ 
& \textbf{106:} \texttt{tiled floor} & - \\ 
& \textbf{107:} \texttt{wooden floor} & - \\ 
& \textbf{108:} \texttt{flower} & - \\ 
& \textbf{109:} \texttt{fog} & - \\ 
& \textbf{110:} \texttt{food} & - \\ 
& \textbf{111:} \texttt{fruit} & - \\ 
& \textbf{112:} \texttt{furniture} & - \\ 
& \textbf{113:} \texttt{grass} & - \\ 
& \textbf{114:} \texttt{gravel} & - \\ 
& \textbf{115:} \texttt{ground} & - \\ 
& \textbf{116:} \texttt{hill} & - \\ 
& \textbf{117:} \texttt{house} & - \\ 
& \textbf{118:} \texttt{leaves} & - \\ 
& \textbf{119:} \texttt{light} & - \\ 
& \textbf{120:} \texttt{mat} & \texttt{door mat} \\ 
& \textbf{121:} \texttt{metal} & \texttt{metal surface, metallic object} \\ 
& \textbf{122:} \texttt{mirror} & - \\ 
& \textbf{123:} \texttt{moss} & \texttt{spores, mosses} \\ 
& \textbf{124:} \texttt{mountain} & - \\ 
& \textbf{125:} \texttt{mud} & - \\ 
& \textbf{126:} \texttt{napkin} & - \\ 
& \textbf{127:} \texttt{net} & - \\ 
& \textbf{128:} \texttt{paper} & - \\ 
& \textbf{129:} \texttt{pavement} & \texttt{sidewalk, footpath} \\ 
& \textbf{130:} \texttt{pillow} & - \\ 
& \textbf{131:} \texttt{plant} & \texttt{plants} \\ 
& \textbf{132:} \texttt{plastic} & - \\ 
& \textbf{133:} \texttt{platform} & - \\ 
& \textbf{134:} \texttt{playing field} & \texttt{playground} \\ 
& \textbf{135:} \texttt{railing} & - \\ 
& \textbf{136:} \texttt{railroad} & - \\ 
& \textbf{137:} \texttt{river} & - \\ 
& \textbf{138:} \texttt{road} & - \\ 
& \textbf{139:} \texttt{rock} & - \\ 
& \textbf{140:} \texttt{roof} & - \\ 
& \textbf{141:} \texttt{rug} & - \\ 
& \textbf{142:} \texttt{salad} & - \\ 
& \textbf{143:} \texttt{sand} & - \\ 
& \textbf{144:} \texttt{sea} & - \\ 
& \textbf{145:} \texttt{shelf} & - \\ 
& \textbf{146:} \texttt{sky} & - \\ 
& \textbf{147:} \texttt{skyscaper} & - \\ 
& \textbf{148:} \texttt{snow} & - \\ 
\multirow{22}{*}{\rotatebox{90}{\centering Stuffs Classes}}  & \textbf{149:} \texttt{solid material} & - \\ 
& \textbf{150:} \texttt{stairs} & - \\ 
& \textbf{151:} \texttt{stone} & - \\ 
& \textbf{152:} \texttt{straw} & - \\ 
& \textbf{153:} \texttt{structure} & - \\ 
& \textbf{154:} \texttt{table} & - \\ 
& \textbf{155:} \texttt{tent} & - \\ 
& \textbf{156:} \texttt{textile} & - \\ 
& \textbf{157:} \texttt{towel} & - \\ 
& \textbf{158:} \texttt{tree} & - \\ 
& \textbf{159:} \texttt{vegetable} & - \\ 
& \textbf{160:} \texttt{brick wall} & - \\ 
& \textbf{161:} \texttt{concrete wall} & - \\ 
& \textbf{162:} \texttt{wall} & - \\ 
& \textbf{163:} \texttt{wall panel} & - \\ 
& \textbf{164:} \texttt{stone wall} & - \\ 
& \textbf{165:} \texttt{tiled wall} & - \\ 
& \textbf{166:} \texttt{wooden wall} & - \\ 
& \textbf{167:} \texttt{water} & - \\ 
& \textbf{168:} \texttt{drops of water} & - \\ 
& \textbf{169:} \texttt{window blind} & - \\ 
& \textbf{170:} \texttt{window} & - \\ 
& \textbf{171:} \texttt{wood} & \texttt{timber} \\ 
\end{longtable}

\twocolumn

%% file: Images/plot_infoscore_supplablation.tex
\begin{tikzpicture}
\begin{axis} [
     title={},
     width=\textwidth,
     height=.2\textheight,
     xlabel={\footnotesize \textbf{Layer paired with top-1 (Layer3)}},
     ylabel={\footnotesize \tikz \fill[blue] (0,0) rectangle (0.02,0.1); \textbf{mIoU}},
     bar width = 8pt,
     ybar = .05cm,
     xmin=0, xmax=13,
     xtick={1,2,3,4,5,6,7,8,9,10,11},
     xticklabel style = {align=center},
     xticklabels={Layer0, Layer1, Layer, Layer4, Layer5, Layer6, Layer7, Layer8, Layer9, Layer10, Layer11},
     ymin=45.0, ymax=60,
     x tick label style={font=\scriptsize},
     y tick label style={font=\tiny},
     y label style={at={(axis description cs:0.05,.5)},anchor=south},
     x label style={at={(axis description cs:0.5,-.25)},anchor=south},
     ymajorgrids=false,
     xmajorgrids=false,
]

\addplot[color=blue!40, fill=blue!40] coordinates {(1, 57.6) (2, 54.1) (3, 51.6) (4, 53.6) (5, 53.5) (6, 49.0) (7, 53.4) (8, 56.5) (9, 49.8) (10, 55.5) (11, 47.3)};
\end{axis}
\end{tikzpicture}